\documentclass[10pt,twocolumn,letterpaper]{article}

\usepackage[pagenumbers]{cvpr} 

\usepackage{graphicx}
\usepackage{amsmath}
\usepackage{amssymb}
\usepackage{booktabs}
\usepackage{color}
\usepackage{multirow}
\usepackage{algorithm}
\usepackage{algpseudocode}
\usepackage{caption, subcaption}
\usepackage{cite}

\usepackage[pagebackref,breaklinks,colorlinks]{hyperref}

\usepackage[capitalize]{cleveref}
\crefname{section}{Sec.}{Secs.}
\Crefname{section}{Section}{Sections}
\Crefname{table}{Table}{Tables}
\crefname{table}{Tab.}{Tabs.}

\def\cvprPaperID{10640} 
\def\confName{CVPR}
\def\confYear{2023}

\begin{document}

\title{Improving the Transferability of Adversarial Samples by Path-Augmented Method}

\author{Jianping Zhang$ ^{1} $ \qquad
Jen-tse Huang$ ^{1} $ \qquad
Wenxuan Wang$ ^{1} $ \qquad
Yichen Li$ ^{1} $ \qquad
\\
Weibin Wu$ ^{2} $\thanks{Corresponding author.} \qquad
Xiaosen Wang$ ^{3} $ \qquad
Yuxin Su$ ^{2} $ \qquad
Michael R. Lyu$ ^{1} $
\\
$ ^{1} $Department of Computer Science and Engineering, The Chinese University of Hong Kong
\\
$ ^{2} $School of Software Engineering, Sun Yat-sen University
\\
$ ^{3} $Huawei Singular Security Lab, Beijing, China
\\
{\tt\small \{jpzhang, jthuang, wxwang, ycli21, lyu\}@cse.cuhk.edu.hk}
\\
{\tt\small \{wuwb36, suyx35\}@mail.sysu.edu.cn, xiaosen@hust.edu.cn}
}

\maketitle

\begin{abstract}
 Deep neural networks have achieved unprecedented success on diverse vision tasks. However, they are vulnerable to adversarial noise that is imperceptible to humans. This phenomenon negatively affects their deployment in real-world scenarios, especially security-related ones. To evaluate the robustness of a target model in practice, transfer-based attacks craft adversarial samples with a local model and have attracted increasing attention from researchers due to their high efficiency. The state-of-the-art transfer-based attacks are generally based on data augmentation, which typically augments multiple training images from a linear path when learning adversarial samples. However, such methods selected the image augmentation path heuristically and may augment images that are semantics-inconsistent with the target images, which harms the transferability of the generated adversarial samples. To overcome the pitfall, we propose the Path-Augmented Method (PAM). Specifically, PAM first constructs a candidate augmentation path pool. It then settles the employed augmentation paths during adversarial sample generation with greedy search. Furthermore, to avoid augmenting semantics-inconsistent images, we train a Semantics Predictor (SP) to constrain the length of the augmentation path. Extensive experiments confirm that PAM can achieve an improvement of over 4.8\% on average compared with the state-of-the-art baselines in terms of the attack success rates.


\end{abstract}

\section{Introduction}
\label{sec:intro}

Deep neural networks (DNNs) appear to be the state-of-the-art solutions for a wide variety of vision tasks \cite{lin2014microsoft, russakovsky2015imagenet}. However, DNNs are vulnerable to adversarial samples \cite{zhang2022practical}, which are elaborately designed by adding human-imperceptible noise to the clean image to mislead DNNs into wrong predictions. The existence of adversarial samples causes negative effects on security-sensitive DNN-based applications, such as self-driving \cite{pouyanfar2018survey} and medical diagnosis \cite{chen2022tw, chen2022dynamic}. Therefore, it is necessary to understand the DNNs \cite{wu2020towards, Wang2020RethinkingTV, huang2022aeon, Wang2022UnderstandingAI} and enhance attack algorithms to better identify the DNN model's vulnerability, which is the first step to improve their robustness against adversarial samples \cite{liu2022towards, liu2022gradients, wu2019deep}.

\begin{figure}
\small
\centerline{\includegraphics[width=8.0cm]{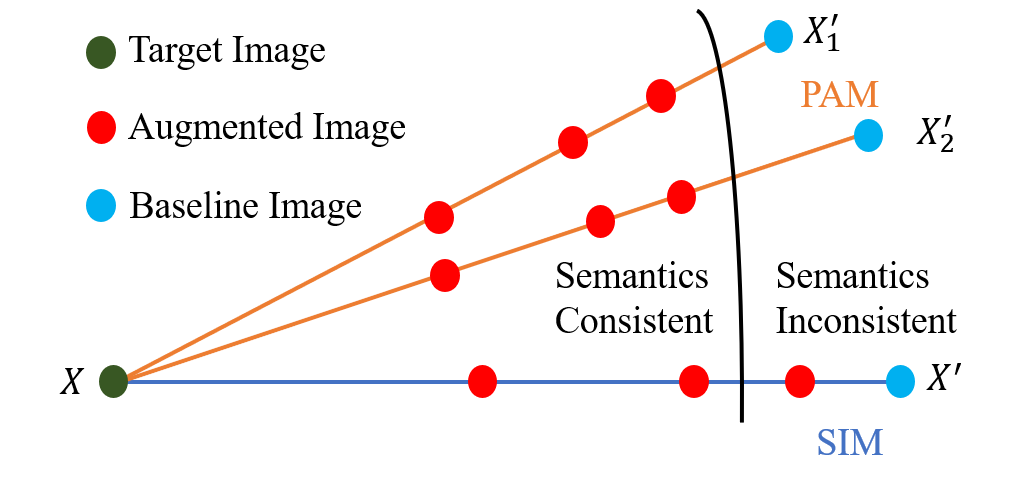}}
\vspace{-3mm}
\caption{
Illustration of how SIM and our PAM augment images (red dots) during the generation of adversarial samples. SIM only considers one linear path from the target image $X$ to a baseline image $X'$. Besides, SIM may augment images that are semantics-inconsistent with the target image. In contrast, our PAM augments images along multiple augmentation paths. We also constrain the length of the path to avoid augmenting images that are semantics-inconsistent with the target one.
}
\label{fig1}
\end{figure}


There are generally two kinds of attacks in the literature \cite{dong2018boosting}. One is the white-box attacks, which consider the white-box setting where attackers can access the architectures and parameters of the victim models. The other is the black-box attacks, which focus on the black-box situation where attackers fail to get access to the specifics of the victim models \cite{wu2020boosting, dong2019evading}. Black-box attacks are more applicable than the white-box counterparts to real-world systems. There are two basic black-box attack methodologies: the query-based \cite{andriushchenko2020square, bai2020improving, wang2022triangle} and the transfer-based attacks \cite{wang2021feature, wu2021improving}. Query-based attacks interact with the victim model to generate adversarial samples, but they may incur excessive queries. In contrast, transfer-based attacks craft adversarial samples with a local source model and do not need to query the victim model. Therefore, transfer-based attacks have attracted more attention recently because of their high efficiency \cite{wu2020boosting, dong2019evading}.


However, transfer-based attacks generally craft adversarial samples by employing white-box strategies like the Fast Gradient Sign Method (FGSM) \cite{goodfellow2014explaining} to attack a local model, which often leads to limited transferability due to overfitting to the employed local model. Most existing solutions address the overfitting issue from the perspective of optimization and generalization, which regards the local model and the target image as the training data of the adversarial sample. Therefore, the transferability of the learned adversarial sample corresponds to its generalization ability across attacking different models \cite{lin2019nesterov}. Such methodologies to improve adversarial transferability can be categorized into two groups. One is the optimizer-based approaches \cite{wang2021enhancing, lin2019nesterov, dong2018boosting, wang2021boosting}, which adopt more advanced optimizers to escape from poor local optima during the generation of adversarial samples. The other is the augmentation-based methods \cite{lin2019nesterov, xie2019improving, dong2019evading, wang2021admix}, which resort to data augmentation and exploit multiple training images to learn a more transferable adversarial sample. 

Current state-of-the-art augmentation-based attacks generally apply a heuristics-based augmentation method. For example, the Scale-Invariant attack Method (SIM) \cite{lin2019nesterov} augments multiple scale copies of the target image, while Admix \cite{wang2021admix} augments multiple scale copies of the mixtures of the target image and the images from other categories. SIM exponentially augments images along a linear path from the target image to a baseline image, which is the origin. Admix, in contrast, first augments the target image with the mixture of the target image and the images from other categories. Then it also exponentially augments images along a linear path from the mixture image to the origin. Therefore, such methods only consider the image augmentation path to one baseline image, i.e., the origin. Besides, although they attempt to augment images that are semantics-consistent to the target image \cite{lin2019nesterov,wang2021admix}, they fail to constrain the length of the image augmentation path, which may result in augmenting semantics-inconsistent images. 

To overcome the pitfalls of existing augmentation-based attacks, we propose a transfer-based attack called Path-Augmented Method (PAM). PAM proposes to augment images from multiple image augmentation paths to improve the transferability of the learned adversarial sample. However, due to the continuous space of images, the possible image augmentation paths starting from the target image are countless. In order to cope with the efficiency problem, we first select representative path directions to construct a candidate augmentation path pool. Then we settle the employed augmentation paths during adversarial sample generation with greedy search. Furthermore, to avoid augmenting semantics-inconsistent images, we train a Semantics Predictor, which is a lightweight neural network, to constrain the length of each augmentation path. 

The difference between our PAM and SIM is illustrated in Figure \ref{fig1}. During the generation of adversarial samples, PAM augments images along multiple image augmentation paths from the target image to different baseline images, while SIM only augments images along a single image augmentation path from the target image to the origin. Besides, PAM constrains the length of the image augmentation path to avoid augmenting images that are far away from the target image and preserve the semantic meaning of the target image. In contrast, SIM may augment images that are semantics-inconsistent with the target image due to the overlong image augmentation path. 

To confirm the superiority of our PAM, we conduct extensive experiments against both undefended and defended models on the ImageNet dataset. Experimental results show that our PAM can achieve an improvement of over 3.7\% on average compared with the state-of-the-art baselines in terms of the attack success rates. We also evaluate the performance of the combination of PAM with other compatible attack methods. Again, experimental results confirm that our method can significantly outperform the state-of-the-art baselines by about 7.2\% on average.

In summary, our contributions in this paper are threefold:
\begin{itemize}
    \item We discover that the state-of-the-art augmentation-based attacks (SIM and Admix) actually augment training images from a linear path for learning adversarial samples. We argue that they suffer from limited and overlong augmentation paths.
    
    \item To address their pitfalls, We propose the Path-Augmented Method (PAM). PAM augments images from multiple augmentation paths during the generation of adversarial samples. Besides, to make the augmented images preserve the semantic meaning of the target image, we train a Semantics Predictor (SP) to constrain the length of each augmentation path.

    \item We conduct extensive experiments to validate the effectiveness of our methodologies. Experimental results confirm that our approaches can outperform the state-of-the-art baselines by a margin of over 3.7\% on average. Besides, when combined with other compatible strategies, our method can significantly surpass the state-of-the-art baselines by 7.2\% on average.
\end{itemize}

\section{Related Work}

\subsection{Adversarial Attack Method}

According to the knowledge of the attacker, there are two categories of attacks in general: white-box and black-box attacks \cite{biggio2018wild}. White-box attacks assume the white-box setting, where attackers have full access to the victim model, including the model structures and parameters. Fast Gradient Sign Method (FGSM) \cite{goodfellow2014explaining} is the first white-box attack that utilizes the sign of the input gradient to maximize the classification loss to generate adversarial samples in one step. Basic Iterative Method (BIM) \cite{kurakin2016adversarial} deploys FGSM to iteratively perturb images to improve the attack performance. Project Gradient Descent (PGD) \cite{madry2017towards} extends BIM with random start to generate diverse adversarial samples. Current white-box attacks can achieve nearly 100\% attack success rates in white-box settings. However, they cannot handle black-box situations, where the model structures and parameters are unseen.

As a result, black-box attacks have attracted increasing attention from researchers recently, which can work in the black-box setting. There are generally two categories of black-box attacks. One is the query-based attacks \cite{papernot2017practical, bhagoji2018practical, guo2019simple, bai2020improving}, and the other is the transfer-based attacks \cite{dong2018boosting, zhang2022improving}. Query-based attacks generally determine the susceptible direction of the victim model through querying it with deliberately designed inputs  \cite{guo2019simple,bhagoji2018practical,bai2020improving}. However, query-based attacks may incur prohibitive query costs, hindering their practical application. Transfer-based attacks exploit the transferability of adversarial samples, which means that the adversarial samples generated by a local source model can also mislead a different target model. Due to their high efficiency, transfer-based attacks are a research hot spot. However, adversarial samples crafted by white-box attacks generally possess limited transferability. 

There are mainly two methodologies to improve the transferability of white-box attacks. The first one is the optimizer-based approach, which aims to escape from poor local optima by adjusting the employment of vanilla gradients during the generation of adversarial samples. For example, Momentum Iterative Fast Gradient Sign Method (MI-FGSM) \cite{dong2018boosting} integrates the momentum term into BIM to improve its adversarial transferability.

The other one is the augmentation-based method, which can be further categorized into two lines. The first one actually augments images from a linear path. For example, Scale Invariant Method (SIM) \cite{lin2019nesterov} exponentially augments images along the linear path from the target image to the origin. Admix \cite{wang2021admix} follows a similar image augmentation path while modifying the starting points as the mixture of the target image and the images from other classes. The other line banks on affine transformations to augment images. For example, the Diverse Input Method (DIM) \cite{xie2019improving} applies random resizing and padding, while Translation Invariant Method (TIM) \cite{dong2019evading} employs shifting. Since affine transformations focus on changing the pixel positions of an image, the augmented images are less diverse than those from a linear path, leading to inferior transferability \cite{wang2021admix}.


Unfortunately, state-of-the-art augmentation-based attacks, like SIM and Admix,  only consider the image augmentation path to one baseline image, i.e., the origin. Besides, they fail to constrain the length of the image augmentation path, which may be overlong and result in augmenting images that are far away from and semantics-inconsistent with the target image. To overcome the deficiencies of such augmentation-based attacks, we propose the Path-Augmented Method (PAM). To make the augmented images more diverse, we propose to augment images from multiple augmentation paths during the generation of adversarial samples. Besides, to make the augmented images preserve the semantic meaning of the target image, we train a Semantics Predictor (SP) to constrain the length of each augmentation path.  As a result, our scheme can achieve superior performance over state-of-the-art transfer-based attacks.

\subsection{Adversarial Defense}

Many adversarial defense methods have been proposed to alleviate the threat of adversarial samples, which can be generally grouped into two categories. The first category is adversarial training, which keeps the state-of-the-art defense methods \cite{tramer2017ensemble, kurakin2016adversarial}. Adversarial training retrains the model by injecting the adversarial samples into the training data to improve its robustness \cite{goodfellow2014explaining}. Ensemble adversarial training augments the training data with perturbations transferred from several other models to defend against transfer-based attacks \cite{kurakin2016adversarial}. The other category is to purify the adversarial samples. They rectify adversarial perturbations by pre-processing inputs without losing classification performance on benign images. The state-of-the-art defense methods in this category include utilizing a high-level representation guided denoiser \cite{liao2018defense}, random resizing and padding \cite{xie2017mitigating}, a JPEG-based defensive compression framework \cite{liu2019feature}, a compression module \cite{jia2019comdefend}, and randomized smoothing \cite{cohen2019certified}. In this paper, we exploit these state-of-the-art defenses to evaluate the effectiveness of our attack against defended models.

\section{Method}

In this section, we first describe the state-of-the-art augmentation-based attacks (SIM and Admix). Then we analyze the limitation of such approaches. We finally present our Path-Augmented Method (PAM) to overcome the pitfalls of such attacks.

\subsection{Augmentation-based Attacks}


We first set up some notations. We denote the benign input image as $x$ and the corresponding true label as $y$. We represent the output of a DNN classifier by $f(x)$. $J(x,y)$ stands for the classification loss function of the classifier, which is usually the cross-entropy loss. Given the target image $x$, adversarial attacks aim to find an adversarial sample $x^{adv}$, which can mislead the classifier, i.e., $f(x^{adv}) \neq f(x)$, while it is human-imperceptible, i.e., satisfying the constraint $\left\| x - x^{adv} \right\|_{p} < \epsilon$. $\left\| \cdot \right\|_{p}$ represents the $L_p$ norm, and we focus on the $L_\infty$ norm here to align with previous papers \cite{dong2018boosting, lin2019nesterov}.

Prevailing white-box attacks like FGSM \cite{goodfellow2014explaining} usually craft adversarial samples by solving the following constrained maximization problem:
\begin{equation} \label{eq1}
\nonumber
    \max_{x^{adv}} \ \ J(x^{adv},y) \ \ \ \ s.t. \left\| x - x^{adv} \right\|_{\infty} < \epsilon.
\end{equation}

\textbf{Scale Invariant Method (SIM)} first computes the average gradient $\bar{g} $ of the classification loss with respect to $m$ scaled copies of the target image. Then it updates the target image with the sign of $\bar{g} $ by a small step size $\epsilon' = \frac{\epsilon}{T}$ in each iteration, where $T$ is the iteration number. The update rule is formulated below:

\begin{equation} \label{eq5}
\begin{split}
    \bar{g}_{t+1} & = \frac{1}{m} \sum_{i=0}^{m-1} \nabla_{x^{adv}_{t}} J( \frac{1}{2^{i}} \cdot  x^{adv}_{t},y), \\   
    x^{adv}_{t+1} & = x^{adv}_t + \epsilon'\cdot sgn\{ \bar{g}_{t+1} \}.   
\end{split}
\end{equation}

\textbf{Admix} first replaces the target image with $m_2$ mixtures of the target image and the images from other categories ($ x' \in X' $). Then it follows SIM by using $m_1$ scale copies of the mixed images. Therefore, Admix computes the update gradient as follows: 

\begin{equation} \label{eq6}
\nonumber
\begin{split}
    &\bar{g}_{t+1}  =  \\
    &\frac{1}{m_1\cdot m_2} \sum_{x' \in X'}\sum_{i=0}^{m_1-1} \nabla_{x^{adv}_{t}} J(\frac{1}{2^{i}} \cdot (x^{adv}_{t}+\eta \cdot(x')),y), \\
\end{split}
\end{equation}
where $ \eta $ is the strength of $ x' $ in the mixture image.  

\subsection{Analysis}

After pre-processing, the pixel value of an image will be normalized. We denote the image with pixel values all equal to 0 as the origin $\mathbf{0}$ in the normalized space. We note that the origin is a pure color image, since all its pixels have constant RGB values when we transform the origin in the normalized space back to the original color space.

We find that when generating adversarial samples, SIM and Admix actually augment images from a linear path. Specifically, SIM augments multiple scaled copies of the target image: $\frac{1}{2^{i}} \cdot  x^{adv}_{t} = \frac{1}{2^i} \cdot x^{adv}_{t} + (1-\frac{1}{2^i}) \cdot \mathbf{0}$, which is a linear combination of the target image and the origin. Therefore, SIM exponentially augments images along a linear path from the target image to the origin. Admix first replaces the target image with the mixture of the target image and the image from other categories ($ x' \in X' $): $x^{adv}_{t}+\eta \cdot x'$. Then it follows SIM to augment multiple scaled copies of the mixture image: $\frac{1}{2^{i}} \cdot  (x^{adv}_{t}+\eta \cdot x') = \frac{1}{2^i} \cdot (x^{adv}_{t}+\eta \cdot x') + (1-\frac{1}{2^i}) \cdot \mathbf{0}$, which is also a linear combination of the mixed target image and the origin. Therefore, Admix exponentially augments images along a linear path from the mixed target image to the origin.

From the above analysis, we argue that SIM and Admix suffer from two pitfalls. The first one is the limited augmentation path. SIM and Admix only consider the augmentation path to one baseline image, which is the origin. However, there are other possible augmentation paths that can increase the diversity of the augmented images. Therefore, the limited diversity of the augmented images can incur limited transferability of the resultant adversarial sample. Besides, the augmentation path of SIM and Admix may be overlong. They may augment images that are too far away from the target image. As a result, the augmented images are close to the origin, which contains no information about the target image. Augmenting such images can distract the learning of adversarial samples against the target image, thus harming adversarial transferability.

\subsection{Path-Augmented Method}
To overcome the pitfalls of state-of-the-art augmentation-based attacks, we propose the Path-Augmented Method (PAM). We first describe how we explore more augmentation paths to increase the diversity of augmented images. Then we introduce our method to constrain the length of the augmentation path to make the augmented images preserve the semantic meanings of the target image.


\subsubsection{Augmentation Path Exploration}

In order to diversify the augmented images, we propose to explore more augmentation paths. In fact, the augmentation paths starting from the target image are numerous, considering the continuous image space. In order to deal with the efficiency problem, we first construct a candidate augmentation path pool by selecting representative augmentation paths. Then, we employ the augmentation path candidate in a greedy manner when crafting adversarial samples.

We first demonstrate the construction of the candidate augmentation path pool. To reduce the numerous searching space and align with SIM, we only consider the pure color images as the baseline image for the augmentation path. Moreover, we select distinct baseline images to guarantee the augmented images on the paths are diverse. The close augmented images have similar augmented gradients having similar effects on transferability. Therefore, we divide the whole image space into multiple regions and select one baseline from each region as the representative augmentation path to form a candidate augmentation path pool. In general, we regard the image space is normalized to [-1, 1] for the RGB channel. We divide each channel by three points (-1, 0, 1) to largely diversify the path, so we have $3^3=27$ representative augmentation paths for the image space. Although we can divide each channel more precisely, the number of augmentation paths increase in cubic degree. Therefore, our way of constructing the augmentation path pool is efficient in improving the transferability.

Afterward, we discuss how to utilize the constructed augmentation path pool for generating adversarial samples. Intuitively, we combine more augmentation paths to compute the gradient, the higher transferability we can obtain, but the computation complexity will increase. Thus, we should balance the transferability and the computation complexity. In consequence, the number of augmentation paths $n$ we select is a hyperparameter to tune. After the determination of the augmentation paths number for computing the gradient, we should also figure out the augmentation paths we select from the candidate augmentation path pool. We first rank the augmentation paths in the candidate path pool by deploying the following adversarial attack and measuring the average transferability on a development dataset to rank each augmentation path. For simplicity, we denote the baseline image from the path pool as $x'$. Therefore the $i$-th scaled augmented image along the path from the target image $x$ to the baseline image $x'$ is represented by $\frac{1}{2^i} \cdot x + (1-\frac{1}{2^i}) \cdot x'$. 

\begin{equation} \label{eq7}
\nonumber
\begin{split}
    \bar{g}_{t+1} & = \frac{1}{m} \sum_{i=0}^{m-1} \nabla_{x^{adv}_{t}} J(\frac{1}{2^i} \cdot x_t^{adv} + (1-\frac{1}{2^i}) \cdot x',y) \\
\end{split}
\end{equation}

We follow a greedy manner in that we choose the top-n augmentation paths and directly combine the gradient of augmented images from those augmentation paths together for generating adversarial samples.

\subsubsection{Semantics Preservation}

In order to keep the semantics of the augmented images on the augmentation paths consistent with the target image, we can constrain the length of the augmentation path and augment the images in the semantics-consistent part to avoid the overlong path. However, it is hard to directly know the semantics-consistent part of the augmentation path. We can use the prediction of the classifier on the image along the augmentation path to identify the semantics-consistent length. If the augmented image is semantics-consistent, the augmented image should have the same prediction as the target image. Therefore, the semantics-consistent length is actually to find the decision boundary of the target image class along the augmentation path. Thus, we train a Semantics Predictor (SP) to constrain the length of each augmentation path. The SP takes the image as the input and predicts the semantic ratio on each augmentation path. The semantic ratio is represented by a scaling factor $r \in [0,1]$ on each augmentation path. Therefore, we can utilize the semantic ratio to constrain the length of the augmentation path. We augment the gradient in the semantics-consistent length to obtain meaningful gradients. Therefore, the $i$-th scaled image along the augmentation path from the target image $x$ to the baseline image $x'$ with a semantic scaling factor $r$ is represented by $\frac{1}{2^i} \cdot x + (1-\frac{1}{2^i})r \cdot x'$.

The Semantics Predictor (SP) is a lightweight neural network consisting of five layers: two Convolutional layers, two Average Pooling layers, and one Fully Connected layer. The image is fed into one Convolutional layer with a kernel size of $5 \times 5$ and one Average Pooling layer with a stride of $4$, which can largely reduce the dimension. Then the feature map is sent into another Convolutional layer and Average Pooling layer with the same setting. After that, the feature map is fed into a Fully Connected layer with Sigmoid activation, and the output size is set to be the number of augmentation paths. The output of the lightweight neural network is exactly the semantic scaling factor of each augmentation path. The training objective is to minimize the difference between the confidence score of the true label and the highest confidence score from other classes, as shown below. We train the Semantics Predictor with Adam optimizer for fifteen epochs and set the learning rate to be $1 \times 10^{-4}$.

\begin{equation} \label{eq8}
\nonumber
\begin{split}
    x_b & = SP(x) \cdot x' + (1-SP(x)) \cdot x\\
    loss & = \left\| F(x_b, y) - \max_{y' \neq y}F(x_b, y') \right\|_{2}
\end{split}
\end{equation}

\subsubsection{Attacking Equation and Comparison}

The attacking equation of PAM is shown below, where $x'_{j}$ is the baseline image of $j$-th augmentation path in the augmentation path pool, and $r_j$ is the semantic ratio of $j$-th augmentation path from the Semantic Predictor. $n$ is the number of augmentation paths, and $m$ is the number of copies. 

\begin{equation} \label{eq9}
\nonumber
\begin{split}
    x^{i,j}_t & =\frac{1}{2^i} \cdot x^{adv}_t + (1-\frac{1}{2^i})r_j \cdot x'_{j} \\
    \bar{g}_{t+1} & = \frac{1}{m\cdot n} \sum_{j=0}^{n-1} \sum_{i=0}^{m-1} \nabla_{x^{adv}_{t}} J(x^{i,j}_t,y) 
\end{split}
\end{equation}

Finally, we regard the current state-of-the-art methods SIM \cite{lin2019nesterov}, and Admix \cite{wang2021admix} are special cases of the PAM because both SIM and Admix treat the origin as the baseline and augment the gradient along a linear path. SIM utilizes the target image as the starting point, but Admixs select mixtures of the target image with images from other classes as starting points. Our PAM tries to solve two problems of the previous methods: the limited and overlong augmentation path. We first augment images from multiple augmentation paths to explore other augmentation directions. Besides, we train a lightweight neural network Semantic Predictor to constrain the length of each augmentation path for providing a semantics-consistent gradient.

\section{Experiments}

In this section, we conduct experiments to validate the effectiveness of our proposed approach. We first specify the setup of the experiments. Then, we present the attacking results of our approach against both state-of-the-art undefended and defended models. Finally, we present the ablation study on the number of augmentation paths and the Semantic Predictor.

\begin{table*}
\small
\centering
\setlength{\tabcolsep}{0.3mm}{
\begin{tabular}{|c|c|ccccccc|} 
\hline
Model & Attack & Inc-v3 & Inc-v4 & IncRes-v2 & Res-v2 & Inc-v3$_{\text{ens3}}$ & Inc-v3$_{\text{ens4}}$ & IncRes-v2$_{\text{adv}}$ \\ 
\hline
\multirow{5}{*}{Inc-v3} & MI-FGSM & \textbf{100.0} & 44.1 & 43.1 & 35.1&  13.2 & 13.2 & 6.2  \\
 & SIM & \textbf{100.0} & 69.9 & 67.7 & 63.2& 36.7  & 31.4 & 17.5  \\
 & VMI & \textbf{100.0} & 71.7 & 67.1 & 59.9 & 36.3  & 31.0 & 17.8 \\
 & Admix & \textbf{100.0} & 80.1 & 79.1 & 70.1& 36.9 & 34.8 & 19.0 \\
 & PAM & \textbf{100.0} & \textbf{83.7} & \textbf{81.2} & \textbf{77.5}  & \textbf{44.8} & \textbf{43.4} & \textbf{22.4}  \\ 
\hline
\multirow{5}{*}{Inc-v4} & MI-FGSM & 55.1 & 99.6 & 46.7 & 41.6& 16.1 & 15.0 & 7.8  \\
 & SIM & 81.2 & 99.5 & 73.8 & 68.7& 47.2 & 44.6 & 29.1  \\
 & VMI & 77.9 & 99.7 & 71.1 & 61.8 & 38.4 & 36.5 & 24.0 \\
 & Admix & 87.0 & 99.7 & 82.9 & 78.2 & 50.6 & 47.5&	31.3 \\
 & PAM & \bf 89.5 & \bf 100.0 & \bf 84.5 & \bf 80.5 & \bf 57.3 & \bf 54.5 & \bf 34.7  \\ 
\hline
\multirow{5}{*}{IncRes-v2} & MI-FGSM & 60.1 & 51.2 & 97.9 & 46.7&  21.0 & 16.0 & 10.9  \\
 & SIM & 84.4 & 80.7 & 99.0 & 76.0 & 56.1 & 48.6 & 41.9  \\
 & VMI & 78.6 & 73.4 & 98.2 & 67.6 &  48.4 & 39.9 & 33.5 \\
 & Admix & 87.7 & 85.3 & 99.1 & 80.4 &	61.4& 54.6 & 47.3 \\
 & PAM & \bf 91.8 & \bf 89.4 & \bf 99.6 & \bf 84.7&  \bf 69.8 & \bf 62.7 & \bf 55.2  \\		
\hline
\multirow{5}{*}{Res-v2} & MI-FGSM & 57.2 & 51.4 & 48.7 & 99.2&  24.2 & 22.4 & 12.7  \\
 & SIM & 74.2 & 70.4 & 68.9 & 99.8& 42.9 & 38.6 & 25.2  \\
 & VMI & 75.0 & 68.8 & 69.4 & 99.3 & 45.6 & 41.0 & 29.6 \\
 & Admix & 80.3& 75.6& 76.1& 99.8&45.5&40.8&	27.5 \\
 & PAM & \bf 81.8 & \bf 77.4 & \bf 76.9 & \bf 100.0& \bf 53.1 & \bf 47.0 & \bf 33.2  \\
\hline
\end{tabular}}
\vspace{-2mm}
\caption{The attack success rates (\%) against seven models by various transfer-based attacks. The best results are marked in bold.}
\label{table1}
\vspace{-2mm}
\end{table*}

\subsection{Experimental Setup}

We focus on attacking image classification models trained on ImageNet \cite{russakovsky2015imagenet}, which is the most widely recognized benchmark task for transfer-based attacks \cite{carlini2019evaluating, kurakin2018adversarial, wu2021improving} and is a more challenging dataset compared to MNIST and CIFAR-10. We follow the protocol of the baseline method \cite{lin2019nesterov} to set up the experiments, whose details are shown as follows.

\textbf{Dataset}. We randomly sample 1000 images of different categories from the ILSVRC 2012 validation set \cite{russakovsky2015imagenet}. We ensure that nearly all selected test images can be correctly classified by all of the models deployed in this paper. We also randomly sample another 1000 images as the development set to train Semantics Predictor and rank representative augmentation paths.

\begin{table*}
\small
\centering
\setlength{\tabcolsep}{0.3mm}{
\begin{tabular}{|c|c|ccccccc|}
\hline
Model & Attack & Inc-v3 & Inc-v4 & IncRes-v2 & Res-v2 & Inc-v3$_{\text{ens3}}$ & Inc-v3$_{\text{ens4}}$ & IncRes-v2$_{\text{adv}}$ \\ 
\hline
\multirow{4}{*}{Inc-v3} & SIM-DT & 99.0 & 85.7 & 80.3 & 75.1 & 67.6 & 63.1 & 46.0 \\
 & VMI-DT & 99.2 & 78.4 & 75.2 & 67.9 & 58.1 & 57.4 & 44.5 \\
 &Admix-DT&\bf 99.6 & 88.1 & 85.6 & 79.1 & 69.2 & 66.1 & 48.9 \\
 & PAM-DT & 99.4 & \bf 93.4 & \bf 91.5 & \bf 88.4 & \bf 80.5 & \bf 78.6 & \bf 59.8 \\
\hline
\multirow{4}{*}{Inc-v4} & SIM-DT & 86.4 & 98.4 & 84.2 & 77.9& 69.9 & 67.1 & 56.1  \\
 & VMI-DT & 81.4 & 98.4 & 76.4 & 67.0 & 58.8 & 56.7 & 49.8 \\
 & Admix-DT & 88.8&	99.4 & 85.8& 80.2& 72.4& 69.0 & 57.6 \\
 & PAM-DT & \bf 93.9 & \bf 99.7 & \bf 91.5 & \bf 87.2 & \bf 80.1 & \bf 78.1 & \bf 65.2 \\
\hline
\multirow{4}{*}{IncRes-v2} & SIM-DT & 88.2 & 85.6 & 97.4 & 82.2 & 77.6 & 73.2 & 72.7  \\
 & VMI-DT & 78.8 & 77.2 & 94.8 & 71.8 & 63.9 & 59.9 & 59.3 \\
 & Admix-DT & 88.2& 87.4 & 98.2 & 84.0 & 80.0 & 75.4 & 71.8 \\
 & PAM-DT & \bf 95.3 & \bf 93.2 & \bf 99.3 & \bf 90.8& \bf 88.8 & \bf 85.4 & \bf 81.8 \\
\hline
\multirow{4}{*}{Res-v2} & SIM-DT & 85.8 & 80.9 & 84.8 & 98.5& 76.2 & 70.3 & 62.0  \\
 & VMI-DT & 81.0 & 78.8 & 78.3 & 98.1 & 69.5 & 65.7 & 57.2 \\
 & Admix-DT & 89.0& 85.5 & 86.2& \bf 99.9 & 78.2 & 73.1 & 64.5 \\
 & PAM-DT & \bf 90.0 & \bf 86.8 & \bf 88.0 & 99.5&  \bf 84.4 & \bf 80.6 & \bf 71.8 \\
\hline
\end{tabular}}
\vspace{-2mm}
\caption{The attack success rates (\%) on eight models by various transfer-based attacks combined with augmentation-based strategies. The best results are marked in bold.}
\vspace{-2mm}
\label{table2}
\end{table*}

\begin{table*}
\small
\centering
\begin{tabular}{|c|ccccccc|} 
\hline
Attack & HGD & R\&P & NIPS-r3 & FD & ComDefend & RS & Average \\ 
\hline
SIM & 15.1 & 28.1 & 36.6 & 59.5& 55.1 & 22.3 & 36.1 \\
VMI & 15.8 & 27.0 & 33.3 & 54.8 & 52.0 & 22.5 & 34.2 \\
Admix & 32.4 & 30.5 & 41.3 & 64.4 & 60.8 & 23.7 & 42.2 \\
PAM & \bf 41.0 & \bf 40.3 & \bf 48.1 & \bf 66.0 & \bf 63.8 & \bf 24.3 & \bf 47.3 \\
\hline
\end{tabular}

\caption{The attack success rates (\%) of six advanced defense mechanisms on adversarial samples. The adversarial samples are generated on the Inc-v3 model by various transfer-based attacks. The best results are marked in bold.}
\label{table3}

\end{table*}

\textbf{Target Model}. We consider both undefended (normally trained) models and defended models as the target models. For undefended models, we choose four top-performance models with different architectures, containing Inception-v3 (Inc-v3) \cite{szegedy2016rethinking}, Inception-v4 (Inc-v4) \cite{szegedy2017inception}, Inception-Resnet-v2 (IncRes-v2) \cite{szegedy2017inception}, and Resnet-v2-101 (Res-v2) \cite{he2016deep, he2016identity}. For defended models, we consider three adversarially trained models, because adversarial training is the most simple but effective way to defend attacks \cite{madry2017towards, xu2022a2}. The selected defended models include Inception v3 trained with adversarial samples from an ensemble of three models (Inc-v3$_{\text{ens3}}$), and four models (Inc-v3$_{\text{ens4}}$), and adversarially trained Inception-Resnet-v2 (IncRes-v2$_{\text{adv}}$). Furthermore, we include six advanced defense models that are robust against black-box attacks on the ImageNet dataset. These defenses cover high-level representation guided denoiser (HGD) \cite{liao2018defense}, random resizing and padding (R\&P) \cite{xie2017mitigating}, NIPS-r3\footnote{https://github.com/anlthms/nips-2017/tree/master/mmd}, feature distillation (FD) \cite{liu2019feature}, compression defense (ComDefend) \cite{jia2019comdefend}, and randomized smoothing (RS) \cite{cohen2019certified}.

\textbf{Baseline}. We take an advanced optimizer-based attack: MI-FGSM \cite{dong2018boosting} as our baseline because it exhibits better transferability than white-box attacks \cite{goodfellow2014explaining, kurakin2016adversarial}. Furthermore, SIM \cite{lin2019nesterov} and Admix \cite{wang2021admix} can be viewed as special cases of our proposed PAM, so we select them as baselines. In order to show that our approaches achieve state-of-the-art performance, we select Variance Tuning Method \cite{wang2021enhancing} (VMI) because Admix and VMI are the current state-of-the-art transfer-based attack methods. In addition, we integrate all the methods with other augmentation-based methods: DIM \cite{xie2019improving} and TIM \cite{dong2019evading} for further comparison. We denote the approaches with DT extension as the method combined with DIM and TIM.

\textbf{Metric}. We evaluate the performance of attack methods via the attack success rate against the target model. The attack success rate is the percentage of adversarial samples that successfully mislead the target model over the total number of the generated adversarial sample. 

\textbf{Parameter}. Following \cite{dong2018boosting}, we set the maximum perturbation budget $\epsilon = 16$, the number of attack iterations $T = 10$, and the step length $\epsilon' = 1.6$. We set the decay factor $\mu = 1.0$ for all the methods. 
We follow the source code of SIM \cite{lin2019nesterov} and Admix \cite{wang2021admix} to change the number of scale copies to 32 and 8 for a fair comparison with the same computation complexity as PAM. For DIM, we set the transformation probability to 0.5. We deploy the 7 × 7 Gaussian kernels for TIM. We take $n = 8$ and $m = 4$ for PAM.

\subsection{Attack Transferability}

First, we study the performance of our attack method PAM against both undefended and defended models. We fix a source model and produce adversarial samples with different attack methods. The generated samples are then fed into the target models to compute the attack success rates. Our attack achieves nearly 100\% success rates under the white-box scenarios in Table \ref{table1}. More importantly, on the evaluation of transferability, our technique can drastically outperform VMI over 10\% and Admix about 4.8\% under the black-box setting on average. In addition, PAM improves the transferability to adversarially trained models, largely showing a high threat to adversarial training. Besides, our attack consistently outperforms other baselines by a significant margin under the black-box setting, which confirms the superiority of our strategies on transferable adversarial sample generation.

Then, we combine all the baselines with augmentation-based methods: DIM and TIM to further enhance the transferability. As shown in Table \ref{table2}, the attack success rates against black-box models are promoted by a large margin with our approaches. In general, our attacks consistently outperform the state-of-the-art baselines by about 7.2\%, which further corroborates the effectiveness of our method.     

In addition, we also evaluate the performance of different attacks against advanced defenses. Table \ref{table3} shows the results when adopting Inc-v3 as the source model to attack other advanced defense models. Our attacks reduce the accuracy of defended models to 52.7\% on average, defeating all baseline attacks. It validates the effectiveness of our attack against advanced defense models, raising security concerns for developing more robust defenses.

\subsection{Ablation Study}

We conduct ablation studies to examine two designs in our proposed PAM: the number of augmentation paths $n$ and the Semantics Predictor. Adversarial samples are generated by attacking the Inc-v3 model without employing augmentation-based methods.

\begin{figure}[t]
  \small
  \centering
  \centerline{\includegraphics[width=0.95\linewidth]{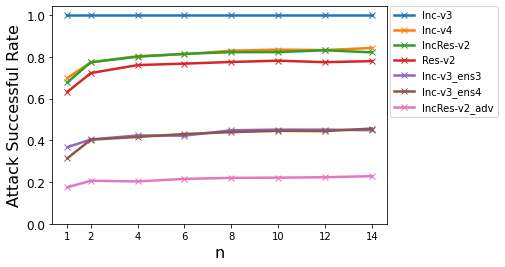}}

  \caption{
  The  attack  success  rates  (\%)  of  PAM with different number of augmentation paths $n$.
  }

  \label{fig2}
\end{figure}

\textbf{Number of Augmentation Paths}. We investigate the effect of different augmentation path numbers on attack performance. We employ PAM with top-$n$ augmentation paths for generating adversarial samples based on the Inc-v3 model. The result is shown in Figure \ref{fig2}. With the increase of the number of augmentation paths, transferability improves. However, the computation cost also rises as the number of augmentation paths increases. Therefore, we choose $n=8$ to balance the performance and computation cost. Besides, we find an intriguing observation that the selected augmentation path is not the same as SIM or Admix when $n = 1$. Our top-1 augmentation path improves the transferability of SIM with more than 1\% on average without introducing additional computation complexity. This means the augmentation path of SIM is not optimal.

\begin{figure}
\small
\centering
\begin{minipage}[t]{0.15\textwidth}
\centering
\includegraphics[width=2.2cm]{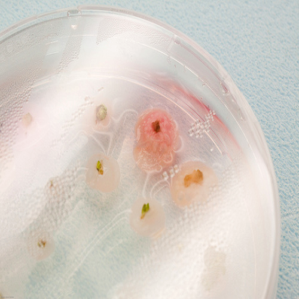}
\subcaption*{Original Image}
\end{minipage}
\begin{minipage}[t]{0.15\textwidth}
\centering
\includegraphics[width=2.2cm]{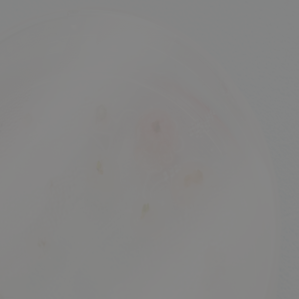}
\subcaption*{SIM}
\end{minipage}
\begin{minipage}[t]{0.15\textwidth}
\centering
\includegraphics[width=2.2cm]{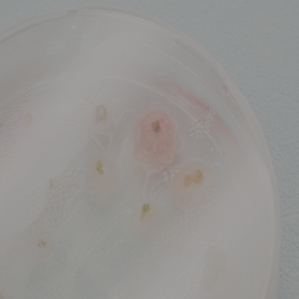}
\subcaption*{SIM+SP}
\end{minipage}

\caption{Visualization of original image and augmented images. We fail to identify the object in the augmented image of SIM. However, the object in the augmented image of SIM+SP is recognizable.}

\label{augmented}
\end{figure}

\textbf{Semantics Predictor}. We study the influence of Semantics Predictor on attack performance for PAM and the performance improvement for SIM. As shown in Table \ref{table4}, the transferability of SIM can be improved by 1\% on average by utilizing the Semantics Predictor because some of the augmented images are semantics-inconsistent with the target image as shown in Figure \ref{augmented}. We cannot recognize the object in the augmented image of SIM. However, the augmented image of SIM+SP demonstrates consistency with the target image, which shows the effectiveness of the Semantic Predictor. In addition, SIM+path+SP outperforms SIM+path by more than 1.6\%, showing Semantics Predictor improves transferability when we combine multiple augmentation paths together. Besides, SIM + path surpasses SIM by a large margin, which also demonstrates the effectiveness of exploring more augmentation paths.
 
\begin{table}
\centering
\small
\setlength{\tabcolsep}{0.3mm}{                           
\begin{tabular}{|l|cccc|} 
\hline
Model & IncRes-v2 & Res-v2 & Inc-v3$_{\text{ens4}}$ & IncRes-v2$_{\text{adv}}$ \\ 
\hline
SIM & 67.7 & 63.2 & 31.4 & 17.5  \\
SIM+SP & 68.3 & 64.3 & 32.5 & 18.3  \\
SIM+paths & 79.7 & 76.4 & 40.7 & 21.2  \\
SIM+paths+SP (PAM) & 81.2 & 77.5 & 43.4 & 22.4  \\
\hline
\end{tabular}}

\caption{The attack success rates (\%) against selected four black-box models by various transfer-based attacks.}
\label{table4}
\end{table}
 
\section{Conclusion}

In this paper, we argue current data augmentation-based attacks suffer from the limited and overlong augmentation path. PAM proposes to augment images from multiple image augmentation paths to improve the transferability of the learned adversarial sample. However, due to the continuous space of images, the possible image augmentation paths starting from the target image are countless. In order to cope with the efficiency problem, we first select representative path directions to construct a candidate augmentation path pool. Then we settle the employed augmentation paths during adversarial sample generation with greedy search. Furthermore, to avoid augmenting semantics-inconsistent images, we train a Semantics Predictor, which is a lightweight neural network, to constrain the length of each augmentation path.  Extensive experiments confirm the superiority of our approaches on generating transferable adversarial samples against both undefended and defended models over state-of-the-art baselines. 

\section*{Acknowledgment}
The work described in this paper was supported by the National Natural Science Foundation of China (Grant No. 62206318) and the Research Grants Council of the Hong Kong Special Administrative Region, China (CUHK 14206921 of the General Research Fund).

{\small
\bibliographystyle{ieee_fullname}
\bibliography{egbib}

\begin{thebibliography}{10}\itemsep=-1pt

\bibitem{andriushchenko2020square}
Maksym Andriushchenko, Francesco Croce, Nicolas Flammarion, and Matthias Hein.
\newblock Square attack: a query-efficient black-box adversarial attack via
  random search.
\newblock In {\em European Conference on Computer Vision}, pages 484--501.
  Springer, 2020.

\bibitem{bai2020improving}
Yang Bai, Yuyuan Zeng, Yong Jiang, Yisen Wang, Shu-Tao Xia, and Weiwei Guo.
\newblock Improving query efficiency of black-box adversarial attack.
\newblock In {\em Computer Vision--ECCV 2020: 16th European Conference,
  Glasgow, UK, August 23--28, 2020, Proceedings, Part XXV 16}, pages 101--116.
  Springer, 2020.

\bibitem{bhagoji2018practical}
Arjun~Nitin Bhagoji, Warren He, Bo Li, and Dawn Song.
\newblock Practical black-box attacks on deep neural networks using efficient
  query mechanisms.
\newblock In {\em Proceedings of the European Conference on Computer Vision
  (ECCV)}, pages 154--169, 2018.

\bibitem{biggio2018wild}
Battista Biggio and Fabio Roli.
\newblock Wild patterns: Ten years after the rise of adversarial machine
  learning.
\newblock {\em Pattern Recognition}, 84:317--331, 2018.

\bibitem{carlini2019evaluating}
Nicholas Carlini, Anish Athalye, Nicolas Papernot, Wieland Brendel, Jonas
  Rauber, Dimitris Tsipras, Ian Goodfellow, Aleksander Madry, and Alexey
  Kurakin.
\newblock On evaluating adversarial robustness.
\newblock {\em arXiv preprint arXiv:1902.06705}, 2019.

\bibitem{chen2022dynamic}
Wenting Chen, Yifan Liu, Jiancong Hu, and Yixuan Yuan.
\newblock Dynamic depth-aware network for endoscopy super-resolution.
\newblock {\em IEEE Journal of Biomedical and Health Informatics},
  26(10):5189--5200, 2022.

\bibitem{chen2022tw}
Wenting Chen, Shuang Yu, Kai Ma, Wei Ji, Cheng Bian, Chunyan Chu, Linlin Shen,
  and Yefeng Zheng.
\newblock Tw-gan: Topology and width aware gan for retinal artery/vein
  classification.
\newblock {\em Medical Image Analysis}, 77:102340, 2022.

\bibitem{cohen2019certified}
Jeremy Cohen, Elan Rosenfeld, and Zico Kolter.
\newblock Certified adversarial robustness via randomized smoothing.
\newblock In {\em International Conference on Machine Learning}, pages
  1310--1320. PMLR, 2019.

\bibitem{dong2018boosting}
Yinpeng Dong, Fangzhou Liao, Tianyu Pang, Hang Su, Jun Zhu, Xiaolin Hu, and
  Jianguo Li.
\newblock Boosting adversarial attacks with momentum.
\newblock In {\em Proceedings of the IEEE conference on computer vision and
  pattern recognition}, pages 9185--9193, 2018.

\bibitem{dong2019evading}
Yinpeng Dong, Tianyu Pang, Hang Su, and Jun Zhu.
\newblock Evading defenses to transferable adversarial examples by
  translation-invariant attacks.
\newblock In {\em Proceedings of the IEEE/CVF Conference on Computer Vision and
  Pattern Recognition}, pages 4312--4321, 2019.

\bibitem{goodfellow2014explaining}
Ian~J Goodfellow, Jonathon Shlens, and Christian Szegedy.
\newblock Explaining and harnessing adversarial examples.
\newblock {\em arXiv preprint arXiv:1412.6572}, 2014.

\bibitem{guo2019simple}
Chuan Guo, Jacob Gardner, Yurong You, Andrew~Gordon Wilson, and Kilian
  Weinberger.
\newblock Simple black-box adversarial attacks.
\newblock In {\em International Conference on Machine Learning}, pages
  2484--2493. PMLR, 2019.

\bibitem{he2016deep}
Kaiming He, Xiangyu Zhang, Shaoqing Ren, and Jian Sun.
\newblock Deep residual learning for image recognition.
\newblock In {\em Proceedings of the IEEE conference on computer vision and
  pattern recognition}, pages 770--778, 2016.

\bibitem{he2016identity}
Kaiming He, Xiangyu Zhang, Shaoqing Ren, and Jian Sun.
\newblock Identity mappings in deep residual networks.
\newblock In {\em European conference on computer vision}, pages 630--645.
  Springer, 2016.

\bibitem{huang2022aeon}
Jen-tse Huang, Jianping Zhang, Wenxuan Wang, Pinjia He, Yuxin Su, and Michael~R
  Lyu.
\newblock Aeon: a method for automatic evaluation of nlp test cases.
\newblock In {\em Proceedings of the 31st ACM SIGSOFT International Symposium
  on Software Testing and Analysis}, pages 202--214, 2022.

\bibitem{jia2019comdefend}
Xiaojun Jia, Xingxing Wei, Xiaochun Cao, and Hassan Foroosh.
\newblock Comdefend: An efficient image compression model to defend adversarial
  examples.
\newblock In {\em Proceedings of the IEEE/CVF Conference on Computer Vision and
  Pattern Recognition}, pages 6084--6092, 2019.

\bibitem{kurakin2018adversarial}
Alexey Kurakin, Ian Goodfellow, Samy Bengio, Yinpeng Dong, Fangzhou Liao, Ming
  Liang, Tianyu Pang, Jun Zhu, Xiaolin Hu, Cihang Xie, et~al.
\newblock Adversarial attacks and defences competition.
\newblock In {\em The NIPS'17 Competition: Building Intelligent Systems}, pages
  195--231. Springer, 2018.

\bibitem{kurakin2016adversarial}
Alexey Kurakin, Ian Goodfellow, Samy Bengio, et~al.
\newblock Adversarial examples in the physical world, 2016.

\bibitem{liao2018defense}
Fangzhou Liao, Ming Liang, Yinpeng Dong, Tianyu Pang, Xiaolin Hu, and Jun Zhu.
\newblock Defense against adversarial attacks using high-level representation
  guided denoiser.
\newblock In {\em Proceedings of the IEEE Conference on Computer Vision and
  Pattern Recognition}, pages 1778--1787, 2018.

\bibitem{lin2019nesterov}
Jiadong Lin, Chuanbiao Song, Kun He, Liwei Wang, and John~E Hopcroft.
\newblock Nesterov accelerated gradient and scale invariance for adversarial
  attacks.
\newblock {\em arXiv preprint arXiv:1908.06281}, 2019.

\bibitem{lin2014microsoft}
Tsung-Yi Lin, Michael Maire, Serge Belongie, James Hays, Pietro Perona, Deva
  Ramanan, Piotr Doll{\'a}r, and C~Lawrence Zitnick.
\newblock Microsoft coco: Common objects in context.
\newblock In {\em European conference on computer vision}, pages 740--755.
  Springer, 2014.

\bibitem{liu2019feature}
Zihao Liu, Qi Liu, Tao Liu, Nuo Xu, Xue Lin, Yanzhi Wang, and Wujie Wen.
\newblock Feature distillation: Dnn-oriented jpeg compression against
  adversarial examples.
\newblock In {\em 2019 IEEE/CVF Conference on Computer Vision and Pattern
  Recognition (CVPR)}, pages 860--868. IEEE, 2019.

\bibitem{liu2022gradients}
Zihan Liu, Yun Luo, Lirong Wu, Siyuan Li, Zicheng Liu, and Stan~Z Li.
\newblock Are gradients on graph structure reliable in gray-box attacks?
\newblock In {\em Proceedings of the 31st ACM International Conference on
  Information \& Knowledge Management}, pages 1360--1368, 2022.

\bibitem{liu2022towards}
Zihan Liu, Yun Luo, Lirong Wu, Zicheng Liu, and Stan~Z Li.
\newblock Towards reasonable budget allocation in untargeted graph structure
  attacks via gradient debias.
\newblock In {\em Advances in Neural Information Processing Systems}, 2022.

\bibitem{madry2017towards}
Aleksander Madry, Aleksandar Makelov, Ludwig Schmidt, Dimitris Tsipras, and
  Adrian Vladu.
\newblock Towards deep learning models resistant to adversarial attacks.
\newblock {\em arXiv preprint arXiv:1706.06083}, 2017.

\bibitem{papernot2017practical}
Nicolas Papernot, Patrick McDaniel, Ian Goodfellow, Somesh Jha, Z~Berkay Celik,
  and Ananthram Swami.
\newblock Practical black-box attacks against machine learning.
\newblock In {\em Proceedings of the 2017 ACM on Asia conference on computer
  and communications security}, pages 506--519, 2017.

\bibitem{pouyanfar2018survey}
Samira Pouyanfar, Saad Sadiq, Yilin Yan, Haiman Tian, Yudong Tao, Maria~Presa
  Reyes, Mei-Ling Shyu, Shu-Ching Chen, and Sundaraja~S Iyengar.
\newblock A survey on deep learning: Algorithms, techniques, and applications.
\newblock {\em ACM Computing Surveys (CSUR)}, 51(5):1--36, 2018.

\bibitem{russakovsky2015imagenet}
Olga Russakovsky, Jia Deng, Hao Su, Jonathan Krause, Sanjeev Satheesh, Sean Ma,
  Zhiheng Huang, Andrej Karpathy, Aditya Khosla, Michael Bernstein, et~al.
\newblock Imagenet large scale visual recognition challenge.
\newblock {\em International journal of computer vision}, 115(3):211--252,
  2015.

\bibitem{szegedy2017inception}
Christian Szegedy, Sergey Ioffe, Vincent Vanhoucke, and Alexander~A Alemi.
\newblock Inception-v4, inception-resnet and the impact of residual connections
  on learning.
\newblock In {\em Thirty-first AAAI conference on artificial intelligence},
  2017.

\bibitem{szegedy2016rethinking}
Christian Szegedy, Vincent Vanhoucke, Sergey Ioffe, Jon Shlens, and Zbigniew
  Wojna.
\newblock Rethinking the inception architecture for computer vision.
\newblock In {\em Proceedings of the IEEE conference on computer vision and
  pattern recognition}, pages 2818--2826, 2016.

\bibitem{tramer2017ensemble}
Florian Tram{\`e}r, Alexey Kurakin, Nicolas Papernot, Ian Goodfellow, Dan
  Boneh, and Patrick McDaniel.
\newblock Ensemble adversarial training: Attacks and defenses.
\newblock {\em arXiv preprint arXiv:1705.07204}, 2017.

\bibitem{Wang2022UnderstandingAI}
Wenxuan Wang, Wenxiang Jiao, Yongchang Hao, Xing Wang, Shuming Shi, Zhaopeng
  Tu, and Michael~R. Lyu.
\newblock Understanding and improving sequence-to-sequence pretraining for
  neural machine translation.
\newblock In {\em Annual Meeting of the Association for Computational
  Linguistics}, 2022.

\bibitem{Wang2020RethinkingTV}
Wenxuan Wang and Zhaopeng Tu.
\newblock Rethinking the value of transformer components.
\newblock In {\em International Conference on Computational Linguistics}, 2020.

\bibitem{wang2021enhancing}
Xiaosen Wang and Kun He.
\newblock Enhancing the transferability of adversarial attacks through variance
  tuning.
\newblock In {\em Proceedings of the IEEE/CVF Conference on Computer Vision and
  Pattern Recognition}, pages 1924--1933, 2021.

\bibitem{wang2021admix}
Xiaosen Wang, Xuanran He, Jingdong Wang, and Kun He.
\newblock Admix: Enhancing the transferability of adversarial attacks.
\newblock In {\em Proceedings of the IEEE/CVF International Conference on
  Computer Vision}, pages 16158--16167, 2021.

\bibitem{wang2021boosting}
Xiaosen Wang, Jiadong Lin, Han Hu, Jingdong Wang, and Kun He.
\newblock {Boosting Adversarial Transferability through Enhanced Momentum}.
\newblock In {\em British Machine Vision Conference}, 2021.

\bibitem{wang2022triangle}
Xiaosen Wang, Zeliang Zhang, Kangheng Tong, Dihong Gong, Kun He, Zhifeng Li,
  and Wei Liu.
\newblock Triangle attack: A query-efficient decision-based adversarial attack.
\newblock In {\em Computer Vision--ECCV 2022: 17th European Conference, Tel
  Aviv, Israel, October 23--27, 2022, Proceedings, Part V}, pages 156--174.
  Springer, 2022.

\bibitem{wang2021feature}
Zhibo Wang, Hengchang Guo, Zhifei Zhang, Wenxin Liu, Zhan Qin, and Kui Ren.
\newblock Feature importance-aware transferable adversarial attacks.
\newblock {\em arXiv preprint arXiv:2107.14185}, 2021.

\bibitem{wu2020boosting}
Weibin Wu, Yuxin Su, Xixian Chen, Shenglin Zhao, Irwin King, Michael~R Lyu, and
  Yu-Wing Tai.
\newblock Boosting the transferability of adversarial samples via attention.
\newblock In {\em Proceedings of the IEEE/CVF Conference on Computer Vision and
  Pattern Recognition}, pages 1161--1170, 2020.

\bibitem{wu2020towards}
Weibin Wu, Yuxin Su, Xixian Chen, Shenglin Zhao, Irwin King, Michael~R Lyu, and
  Yu-Wing Tai.
\newblock Towards global explanations of convolutional neural networks with
  concept attribution.
\newblock In {\em Proceedings of the IEEE/CVF Conference on Computer Vision and
  Pattern Recognition}, pages 8652--8661, 2020.

\bibitem{wu2021improving}
Weibin Wu, Yuxin Su, Michael~R Lyu, and Irwin King.
\newblock Improving the transferability of adversarial samples with adversarial
  transformations.
\newblock In {\em Proceedings of the IEEE/CVF Conference on Computer Vision and
  Pattern Recognition}, pages 9024--9033, 2021.

\bibitem{wu2019deep}
Weibin Wu, Hui Xu, Sanqiang Zhong, Michael~R Lyu, and Irwin King.
\newblock Deep validation: Toward detecting real-world corner cases for deep
  neural networks.
\newblock In {\em 2019 49th Annual IEEE/IFIP International Conference on
  Dependable Systems and Networks (DSN)}, pages 125--137. IEEE, 2019.

\bibitem{xie2017mitigating}
Cihang Xie, Jianyu Wang, Zhishuai Zhang, Zhou Ren, and Alan Yuille.
\newblock Mitigating adversarial effects through randomization.
\newblock {\em arXiv preprint arXiv:1711.01991}, 2017.

\bibitem{xie2019improving}
Cihang Xie, Zhishuai Zhang, Yuyin Zhou, Song Bai, Jianyu Wang, Zhou Ren, and
  Alan~L Yuille.
\newblock Improving transferability of adversarial examples with input
  diversity.
\newblock In {\em Proceedings of the IEEE/CVF Conference on Computer Vision and
  Pattern Recognition}, pages 2730--2739, 2019.

\bibitem{xu2022a2}
Zhuoer Xu, Guanghui Zhu, Changhua Meng, Shiwen Cui, Zhenzhe Ying, Weiqiang
  Wang, Ming Gu, and Yihua Huang.
\newblock A2: Efficient automated attacker for boosting adversarial training.
\newblock {\em arXiv preprint arXiv:2210.03543}, 2022.

\bibitem{zhang2022improving}
Jianping Zhang, Weibin Wu, Jen-tse Huang, Yizhan Huang, Wenxuan Wang, Yuxin Su,
  and Michael~R Lyu.
\newblock Improving adversarial transferability via neuron attribution-based
  attacks.
\newblock In {\em Proceedings of the IEEE/CVF Conference on Computer Vision and
  Pattern Recognition}, pages 14993--15002, 2022.

\bibitem{zhang2022practical}
Qilong Zhang, Chaoning Zhang, Chaoqun Li, Jingkuan Song, Lianli Gao, and
  Heng~Tao Shen.
\newblock Practical no-box adversarial attacks with training-free hybrid image
  transformation.
\newblock {\em arXiv preprint arXiv:2203.04607}, 2022.

\end{thebibliography}
}

\newpage
\appendix
\section{PAM Algorithm}

This section illustrates the PAM algorithm in detail. The algorithm is shown in Algorithm \ref{alg1}. PAM first deploys the Semantic Predictor to predict the semantic ratio for each augmentation path selected from the augmentation path pool to constrain the length. Then, PAM combines the augmented images along all the constrained augmentation paths together to craft adversarial samples.

\begin{algorithm}[h]
\caption{Path-Augmented Method (PAM)}\label{alg1}
\begin{algorithmic}
\Require A classifier $f$ with parameter $\theta$, loss function $J$
\Require A clean example $x$ with ground-truth label $y$
\Require Perturbation budget $\epsilon$, iteration number $T$
\Require Augmentation path pools $X'$
\Require Semantic Predictor $SP(\cdot)$
\Require augmentation path number $n$ and scale copies $m$
\State $\alpha = \frac{\epsilon}{T}$
\State $x_0^{adv} = x$ 
\State $r = SP(x)$ 
\For{$t = 0$ to $ T-1$}
\State $Grad = \textbf{0}$
\For {$j = 1$ to $  n$}
    \For{$i = 1 $ to $  m$}
    \State $x^{i,j}_t =\frac{1}{2^i} \cdot x^{adv}_t + (1-\frac{1}{2^i})r_j \cdot x'_{j}$
    \State $Grad = Grad + \nabla_{x} J(x^{i,j}_t,y;\theta)$
    \EndFor
\EndFor
\State $x_{t+1}^{adv} = x_{t}^{adv} + \alpha \cdot sign(Grad)$

\EndFor
\end{algorithmic}
\end{algorithm}

\end{document}